\documentclass{article}

\usepackage{arxiv}

\usepackage[utf8]{inputenc} % allow utf-8 input
\usepackage[T1]{fontenc}    % use 8-bit T1 fonts
\usepackage{hyperref}       % hyperlinks
\usepackage{url}            % simple URL typesetting
\usepackage{booktabs}       % professional-quality tables
\usepackage{amsfonts}       % blackboard math symbols
\usepackage{nicefrac}       % compact symbols for 1/2, etc.
\usepackage{microtype}      % microtypography
\usepackage{lipsum}

\title{Physics-based polynomial neural networks for one-shot learning of dynamical systems from one or a few samples}

\author{
  Andrei Ivanov\\
  Deutsches Elektronen Synchrotron DESY\\ Notkestrasse 85, 22607, Hamburg, Germany\\
  \texttt{05x.andrey@gmail.com} \\
   \And
  Uwe Iben \\
  Robert Bosch GmbH\\
  Postfach 10 60 50, 70049 Stuttgart, Germany\\
  \texttt{uwe.iben@de.bosch.com } \\
  \And
  Anna Golovkina \\
  Saint Petersburg State University\\
  University Embankment, 7/9, St Petersburg, Russia, 199034\\
  \texttt{a.golovkina@spbu.ru} \\
}

\usepackage{booktabs} % for professional tables
\usepackage{amsmath}
\newcommand{\XX}{\mathbf{X}}

\newcommand{\FF}{\mathbf{F}}
\newcommand{\MM}{\mathcal M}
\usepackage{graphicx}
\usepackage[]{units} 

\begin{document}
\maketitle

\begin{abstract}
This paper discusses an approach for incorporating prior physical knowledge into the neural network to improve data efficiency and the generalization of predictive models. If the dynamics of a system approximately follows a given differential equation, the Taylor mapping method can be used to initialize the weights of a polynomial neural network. This allows the fine-tuning of the model from one training sample of real system dynamics. The paper describes practical results on real experiments with both a simple pendulum and one of the largest worldwide X-ray source. It is demonstrated in practice that the proposed approach allows recovering complex physics from noisy, limited, and partial observations and provides meaningful predictions for previously unseen inputs. The approach mainly targets the learning of physical systems when state-of-the-art models are difficult to apply given  the lack of training data.
\end{abstract}

% keywords can be removed
\keywords{physics-based neural network \and one-shot learning \and dynamical systems}

%-----------------------------
\section{Introduction}
The traditional approach for representing a dynamical system behavior is physical-based models that are derived on conservation laws, e.g. mass, momentum, energy. These laws include an infinity number of data, that are not explicit data but implicit one. To solve real problems, additional approximate models (drag or friction coefficient, boundary conditions, etc.) are used. Such models introduce some level of simplicity to meet the scale-accuracy trade-off. Also, this approach is generally computationally expensive and requires highly accurate numerical solvers. Replacing such physics-based models with high-performance approximate ones can be based on machine learning (ML) approaches for modeling and identification of dynamical systems.

Another very important approach in dynamical system identification are gray box models. They come in various flavors, using black box models to both infer parameters of a system derived from first principles or using them as error models for such systems. This group of methods requires both statistical data and numerical solvers to build the model and estimate its parameters.

Applying machine learning (ML) methods for dynamical systems learning can avoid numerical solvers completely and extract the complex behaviour of the systems, but this generally requires lots of data for training. The model trained with limited observations is highly likely to lead to unsatisfactory performance. Moreover, no guarantee exists that the black-box model can correctly predict system dynamics for completely new and unseen inputs.

Though some studies demonstrate the application of neural networks (NNs) for physical systems learning \cite{Mohajerin,Jia, Koppe, Bieker, Yu}, the described methods require large volumes of measured or simulated data for NN training. The idea of these methods is building surrogate models that can replace physics-based models. Some authors suggest the gray-box models with incorporating NNs into the differential equation to approximate unknown terms. For example, the authors in \cite{Ayed} propose dynamical systems learning from partial observations with ordinary differential equations (ODEs), while the authors in \cite{raissi2017physicsII} add NNs to partial differential equations. A back-propagation technique through an ODE solver is proposed in \cite{ref9} but requires traditional numerical solvers to simulate the dynamics.

By physics-inspired NNs \cite{Thuerey}, authors generally mean either incorporating domain knowledge in the traditional NN architectures or providing additional loss functions for the physical inconsistency of the predictions \cite{PINN}. All these papers use various architectures of NN for model-free systems learning and control \cite{Nagabandi,ChenY}. Moreover, authors do not consider the predictions with inputs that significantly differ from the training data. The physical inconsistency term is estimated only for training data without generalization on unseen inputs.

%--------------------------------------------------------
\begin{figure}
  \centering
  \includegraphics[width=1\textwidth]{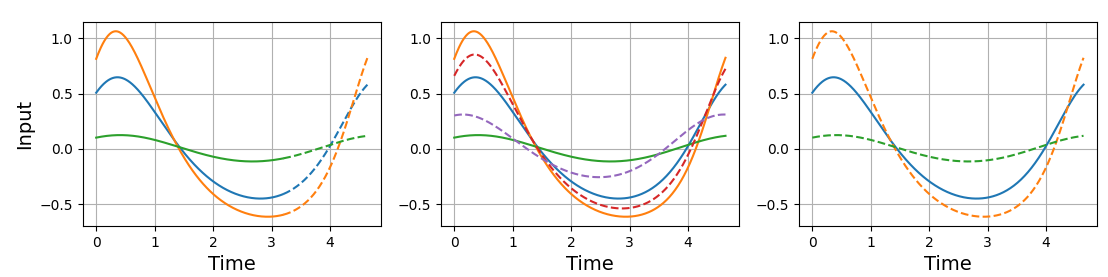}
  \begin{minipage}[t]{0.35\textwidth}
  \centering
  \small{a) forecasting}
  \end{minipage}
  \begin{minipage}[t]{0.33\textwidth}
  \centering
  \small{b) surrogate models}
  \end{minipage}
  \begin{minipage}[t]{0.3\textwidth}
  \centering
  \small{c) one-shot learning}
  \end{minipage}
%   \small{\hspace*{2.0cm}a) forecasting\hspace{2cm}b) surrogate models\hspace{2.0cm}c) one-shot learning}
  \caption{Various problems for learning of dynamical system. Solid lines represent training data, dashed lines corresponds to predictions that are required to be produced by a model.}
  \label{fig.01}
\end{figure}
%--------------------------------------------------------
Traditional and state-of-the-art ML/NN models are suitable for either forecasting or building surrogate models (see Fig.~\ref{fig.01}a,~\ref{fig.01}b). Forecasting means the extrapolation of dynamics in time, while a surrogate model extrapolates dynamics to new inputs that stem from the same distribution as training data. In the paper, we focus on the problems of one-shot learning of the dynamical systems from only one training sample (Fig.~\ref{fig.01}c). The model is requested to predict the dynamics for new inputs beyond the training sample. Since the proposed approach is not an incremental extension of previously studied problems and instead formulates a new problem, we do not compare it with state-of-the-art models or the gray-box approach, which are difficult to apply given the lack of training data.

% In the paper we focus on the problems of one-shot learning of the dynamical systems when only one training sample that describes the system behaviour is available. This means that having measured dynamics for a single input of the system, the model required to predict system behaviour for completely new and unseen inputs. Since the proposed approach is not an incremental extension of previously studied problems but instead formulates a new problem, we do not compare it with state-of-the-art models or gray box approach. For the considered problem, the existing ML and NN models are difficult to apply due to the lack of training data.

To solve the problem of one-shot learning of dynamical systems, we suggest incorporating prior physical knowledge into the NN to improve data efficiency and the generalization of predictive models. We also completely avoid the numerical solvers for dynamical systems learning. The approach presented in the paper is based on \cite{MLM, TMPNN}, where the authors demonstrate how to construct the polynomial neural network (PNN) that approximates the exact system of ODEs and use it for solving differential equations. In contrast to this, we do not target solving exact equations but rely only on an approximate form of the ODEs and the identification of the dynamical system from one training sample.

In the next section, we introduce a brief description of the Taylor mapping approach that is used to translate differential equations into the PNN. Sec.~\ref{sec.3} considers simple examples of free fall and nonlinear oscillation to demonstrate the limitations of existing ML and NN models for learning the simplest dynamical systems when only limited data is available. Sec.~\ref{sec.4} and \ref{sec.5} describes training of the Taylor map-based PNN (TM-PNN) with one sample. The example of one-shot learning of the real pendulum is described in the Sec.~\ref{sec.4} in detail. The same technique is directly adopted to the practical experiments with the largest worldwide X-ray source and the symplectic regularization is introduced in Sec.~\ref{sec.5}  Sec.~\ref{sec.7} briefly describes the application of the discussed methods to other domains where problems of identification of dynamical systems arise widely.

\section{Taylor maps for solving ODEs}
\label{sec.2}
\begin{figure}
\vskip 0.2in
\begin{center}
\centerline{\includegraphics[width=0.4\columnwidth]{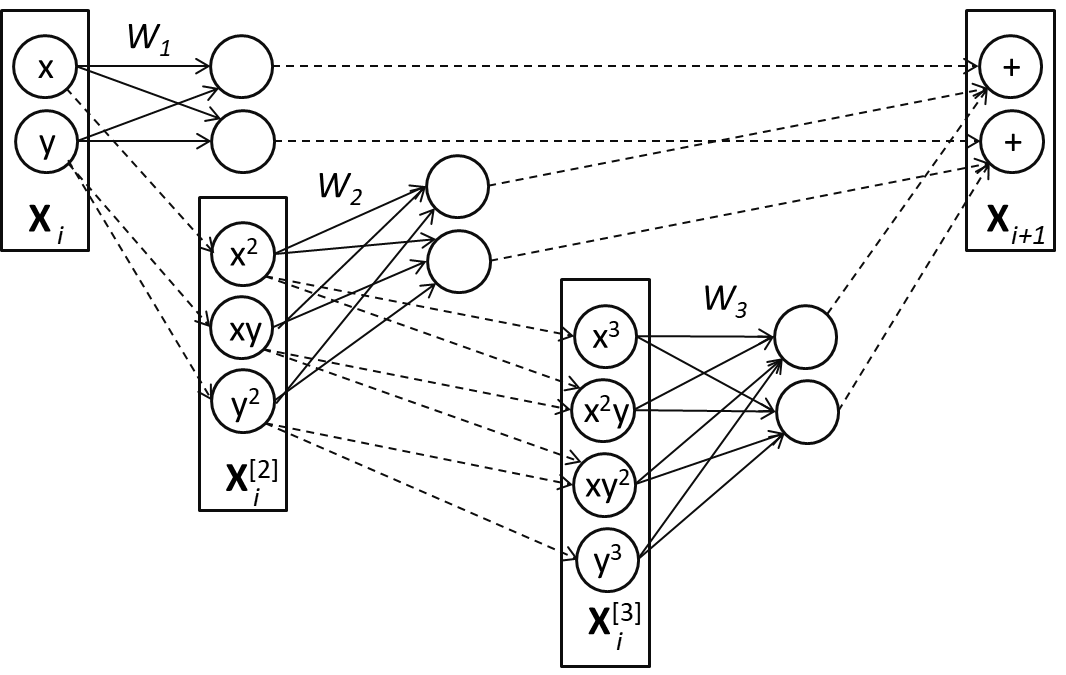}}
\caption{Polynomial neuron of third order nonlinearity.}
\label{fig.02}
\end{center}
\vskip -0.2in
\end{figure}

The transformation $\mathcal M : \XX_0=\XX(t_0)\rightarrow \XX(t_1)$ defines a Taylor map in form of
\begin{equation}
	\label{tmap}
	\XX(t_1) = W_0 + W_1\,\XX_0+W_2\,\XX_0^{[2]}+\ldots+W_k\,\XX_0^{[k]},
\end{equation}
where $\XX, \XX_0 \in R^n$, matrices $W_i$ are weights, and $\XX^{[k]}$ means the $k$th Kronecker power of vector $\XX$ with the same terms reduction. For example, if $\XX = (x_1, x_2)$, then $\XX^{[2]} = (x_1^2, x_1x_2, x_2^2)$ and $\XX^{[3]} = (x_1^3, x_1^2x_2, x_1x_2^2$, $x_2^3)$. The transformation \eqref{tmap} is linear in weights $W_i$ and nonlinear with respect to the $\XX_0$.  In the literature, the transformation \eqref{tmap} can be referred to Taylor maps and models \cite{ref_tm1}, tensor decomposition \cite{ref_tm2}, matrix Lie transform \cite{ref16}, exponential machines \cite{ref201}, and others. In fact, transformation~\eqref{tmap} is just a polynomial regression with respect to the components of $\XX$ that directly defines the polynomial neuron (see Fig.~\ref{fig.02}). 

%--------------------------------------------------------
%--------------------------------------------------------

The Taylor map \eqref{tmap} approximates the general solution of the system of differential equations \cite{ref13,ref15}. If the systems of ODE is known, the weights in \eqref{tmap} can be calculated directly. The initialized from ODEs Taylor map accurately represents the dynamics of the system without the necessity of using numerical solver \cite{TMPNN}.
Indeed, for the differential equation with polynomial right-hand side
%--------------------------------------------------------
\begin{equation}
\label{odesystem}
\frac{d}{dt}\XX = \FF(t, \XX) = \sum_{k=0}^{\infty} P_{k}(t)\XX^{[k]},
\end{equation}
%--------------------------------------------------------
one can find the solution in form of \eqref{tmap} by differentiating it with respect to the $t$
%--------------------------------------------------------
\begin{equation*}
	\frac{d}{dt}\XX(t) = \frac{d}{dt}W_0(t) + \frac{d}{dt}W_1(t)\,\XX_0+\ldots+\frac{d}{dt}W_k(t)\,\XX_0^{[k]},
\end{equation*}
%--------------------------------------------------------
where $t$ is an independent variable, $\XX \in R^n$ is a state vector. The last formula combined with the \eqref{odesystem} yields a new system of ODEs with respect to the weight matrices $W_i$
%--------------------------------------------------------
\begin{equation}
\frac{d}{dt}W_i = f(W_0, \ldots, W_k, P_0, \ldots P_k).
\label{eq:dW}
\end{equation}
%--------------------------------------------------------
where $f_i$ are functions of matrices $W_i$ and $P_i$. For instance, $f_0 = P_0 + P_1W_0 + \ldots P_k W_0^{[k]}$. Solving \eqref{eq:dW} for $W_i$ gives the dynamics of the system for all initial conditions by means of \eqref{tmap}. Since the transformation \eqref{tmap} is assumed to be valid for any initial value $\XX_0$, the last equation for $W_i$ does not depend on $\XX_0$. Moreover, it can be solved only once with the unified initial condition $W_0 = 0, W_1 = I, W_k = 0, k>1$ with $I$ as identity matrix.
% For the demonstration, let us consider the Taylor mapping technique for the simple ODE
% \begin{equation}
%     x' = kx^2,\;\;\;\;x(0) = x0,
%     \label{eq.1}
% \end{equation}
% where $k\in R$. One can find the general solution of \eqref{eq.1} in form of map \eqref{tmap}
% $  x = w_0 + w_1x_0  + w_2x_0^2+ w_3x_0^3 + \ldots $
% by differentiating and and substitution into \eqref{eq.1}:
% \begin{equation}
% \begin{split}
% x' = w'_0 + w'_1x_0  + w'_2x_0^2+ w'_3x_0^3 + \ldots,\;\;\;\;
% x' =k(w_0 + w_1x_0  + w_2x_0^2+ w_3x_0^3 + \ldots)^2.
% \end{split}
% \end{equation}
% These two relations results a new system of ODEs wit respect to the $w_i$:
% \begin{equation}
% \begin{split}
% w'_0 = kw_0,\;\;\;\;
% w'_1 = 2kw_0w_1,\;\;\;\;
% w'_2 = k(2w_0w_2 + w_1^2),\;\;\;\;
% w'_3 = k(2w_0w_3 + 2w_1w_2),\;\ldots\\
% \end{split}
% \end{equation}
% with initial conditions $w_1(0) = 1, w_i(0)=0, i \neq 1$. Solution of this system can be found analytically that results $x(t) = x0 +ktx_0^2 +k^2t^2x_0^3 + \ldots$. Finally, using the geometric sequence one can write $x = x_0/(1-ktx_0).$
%For the given simple example the Taylor mapping approach results an analytical solution, while in general case the numerical estimation of map \eqref{tmap} can be calculated.
A more detailed description of the Taylor mapping approach along with the theoretical estimations of accuracy and convergence of the truncated series \eqref{tmap} for solving systems of ODEs can be found in \cite{ref13,ref15}.

%--------------------------------------------------------
\begin{figure}
\vskip 0.2in
\begin{minipage}[t]{0.5\textwidth}
\begin{center}
\centerline{\includegraphics[width=0.8\columnwidth]{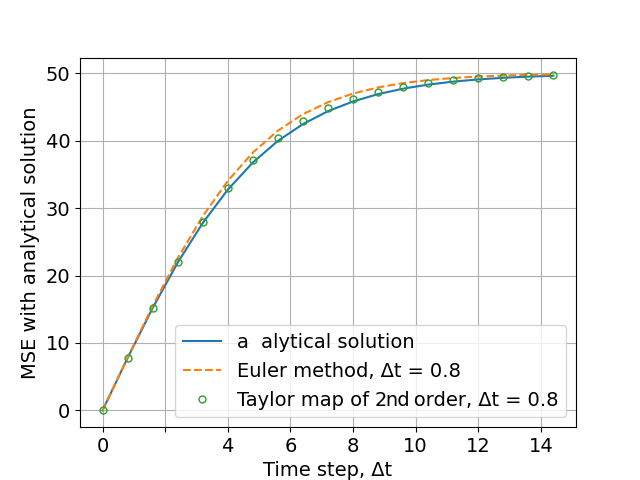}}
\caption{Numerical solutions by Euler method}
and second order Taylor map.
\label{fig.03}
\end{center}
\end{minipage}
\begin{minipage}[t]{0.5\textwidth}
\begin{center}
\centerline{\includegraphics[width=0.8\columnwidth]{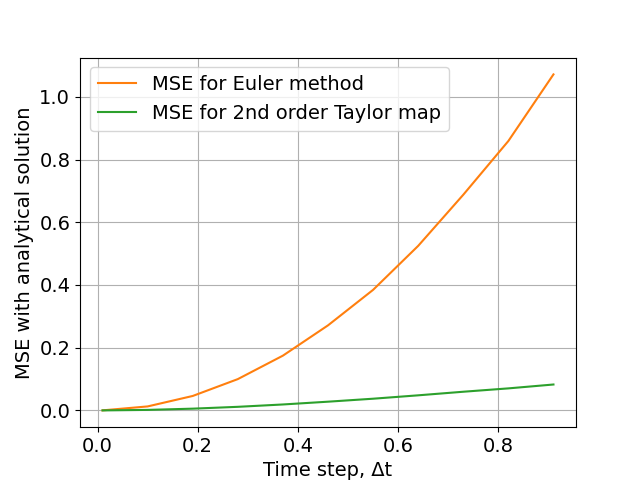}}
\caption{MSE between numerical solutions and analytical one depending on integrating time step.}
\label{fig.04}
\end{center}
\end{minipage}
\vskip -0.2in
\end{figure}
%--------------------------------------------------------
For example, let us consider the model of a body free fall with air resistance
%--------------------------------------------------------
\begin{equation}
v'= g - \frac{k}{m} v^2,
\label{eq.freefall}
\end{equation}
%--------------------------------------------------------
with velocity $v$, mass $m$, the gravitational acceleration $g =\unit[9.8]{kg/m^2}$, resistant coefficient $k=\unit[0.392]{kg/m}$ and $'$ for derivative on time. The traditional approach for solving \eqref{eq.freefall} is using numerical step-by-step solvers. For instance, the Euler method with time step $\Delta t = 0.1$ for $m=\unit[100]{kg}$ and $v_0=v(t)$ results
%--------------------------------------------------------
\begin{equation}
v(t+\Delta t) = g\Delta t + v_0 - kv_0^2\Delta t/m = 0.98 + v_0 - 0.000392v_0^2.
\label{eq.map_euler}
\end{equation}
%--------------------------------------------------------
The Euler method produces a Taylor map \eqref{tmap} of the  second order that is calculated from a first-order discretization of \eqref{eq.freefall}. Using the described in \eqref{tmap}-\eqref{eq:dW} algorithm, one can estimate the second-order Taylor map for \eqref{eq.freefall} more precisely:
%--------------------------------------------------------
\begin{equation}
v(t+\Delta t) = 0.979874527013 + 0.999615938364v_0 -0.0003842685780960v_0^2.
\label{eq.map_tm}
\end{equation}
%--------------------------------------------------------
To compare the accuracy of maps \eqref{eq.map_euler} and \eqref{eq.map_tm} one can use the analytical solution \cite{Lindemuth} of the equation \eqref{eq.freefall} :
%--------------------------------------------------------
\begin{equation}
v(t) = \sqrt{\frac{mg}{k}}\tanh\left(t\sqrt{\frac{kg}{m}}\right).
\label{eq.analytic}
\end{equation}
%--------------------------------------------------------

Fig.~\ref{fig.03} shows the numerical solution of the eq. \eqref{eq.freefall} calculated with the Euler method and Taylor mapping. Both methods result second-order approximation of the exact solution for the time step $\Delta t = \unit[0.8]{s}$. Fig.~\ref{fig.04} presents the mean squared error (MSE) between numerical and analytical solutions for both methods depending on the time step $\Delta t$. One of the Taylor mapping approach advantages is the possibility to solve ODE with a larger step with the necessary level of accuracy in comparison to traditional numerical schemes. For example, the Runge--Kutta method of fourth-order results a 16th order Taylor map for eq. \eqref{eq.freefall}, while a true Taylor map of 16th order can be calculated more accurately for the given equation.

%--------------------------------------------------------
\begin{figure}[ht]
\vskip 0.2in
\begin{center}
\centerline{
\includegraphics[width=0.35\columnwidth]{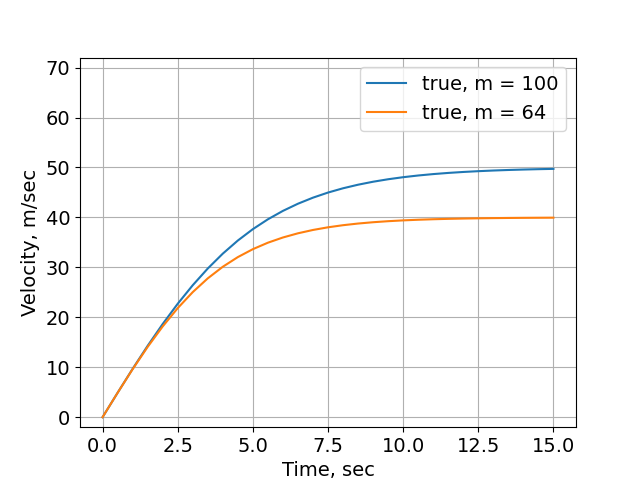}
\includegraphics[width=0.65\columnwidth]{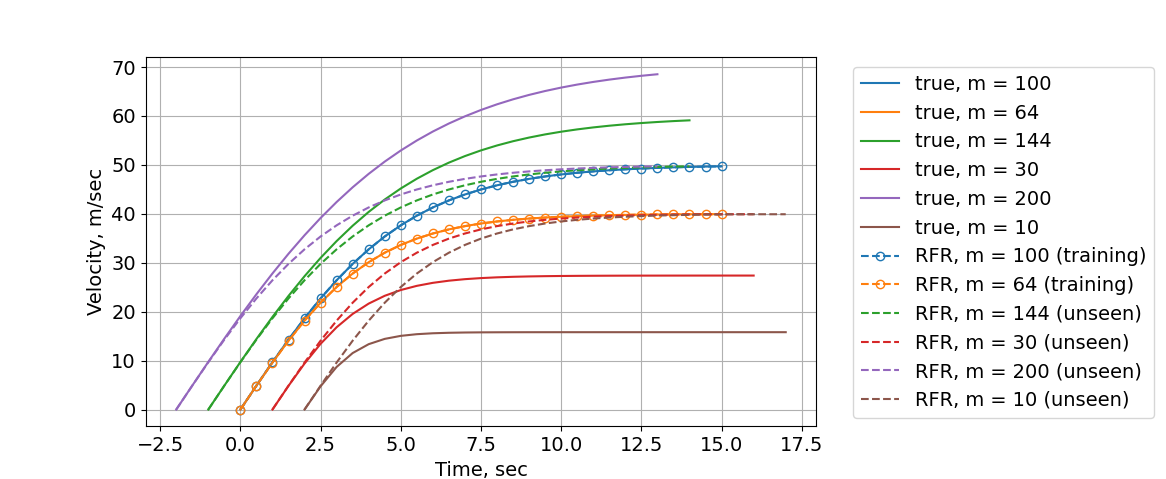}
}
\begin{minipage}[t]{0.3\textwidth}
\centerline{\small{a) training solutions}}
\end{minipage}
\begin{minipage}[t]{0.6\textwidth}
\centerline{\small{b) predictions based on Random Forest regression}}
\end{minipage}
\caption{Left plot shows two solutions for free fall of a body with two different masses as training samples. Right plot demonstrates memorization of the dynamics for a surrogate model based on Random Forest Regressor (RFR) trained with these two solutions.}
\label{fig.05}
\end{center}
\vskip -0.2in
\end{figure}
%--------------------------------------------------------
%--------------------------------------------------------
\begin{figure}[ht]
\vskip 0.2in
\begin{center}
\centerline{\includegraphics[width=0.7\columnwidth]{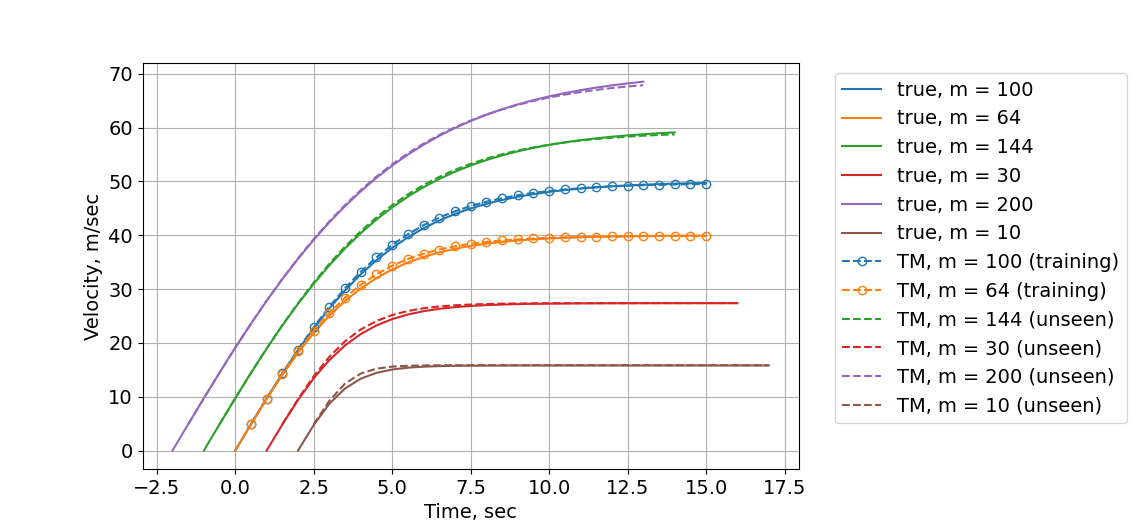}}
\caption{Generalization of the dynamics with a Taylor map trained from data.}
\label{fig.06}
\end{center}
%\vskip -0.2in
\end{figure}
%--------------------------------------------------------
\section{Training TM-PNN from scratch for simple physical systems}
\label{sec.3}
In the paper \cite{TMPNN}, it is demonstrated how to use the Taylor mapping technique to construct PNN that solves the given system of ODEs in various domains. In the rest of the paper, we focus on inverse problems, when the physical laws and equations are not known but the measurements from the system are available. The next section corresponds to examples with virtual measurement data that are generated from simple dynamical models. We use such examples only to demonstrate a problem of one-shot learning of dynamical systems with one or a few training samples.

\subsection{Free fall of a body}
Fig.~\ref{fig.05}a presents two solution of the system \eqref{eq.freefall} with initial velocity $v(0)=0$ and masses $m_1=\unit[100]{kg}$ and $m_2=\unit[80]{kg}$. These solutions are obtained by integrating the equation  with a constant time step $\Delta t = \unit[0.5]{s}$ during time interval $t=\unit[15]{s}$. So, each solution is represented by a univariate time-series $\{v_i\}_{i=0..60}$. Let us use these data for training an ML-based model and validate its prediction for unseen masses. For example, for Random Forest Regression that is commonly used in practice the behavior is presented in Fig.~\ref{fig.05}b. The model accurately predicts the known dynamics but can not represent dynamics for unseen masses. The traditional ML-based model just memorizes these two solutions and attempts to predict them whatever masses are used as inputs. To perform well the ML models require lots of training solutions with different masses. Thus, this questions the ability of such a surrogate model to generalize physics rather than to predict some average behavior based on the presented samples. 

Since the Taylor map \eqref{tmap} corresponds to some ODEs, one can simply estimate the weights of a Taylor map directly from the data and achieve more physical predictions for unseen inputs. Fig.~\ref{fig.06} demonstrates the predictions of a second-order Taylor map that is fitted with the same training solutions:
%--------------------------------------------------------
\begin{equation}
    v(\Delta t=0.5) = 4.864129 + 0.99798733v_0 -0.19208v_0^2/m.
    \label{eq.em2}
\end{equation}
%--------------------------------------------------------
In contrast to traditional ML models, this simple Taylor map can predict dynamics for new masses with a fair accuracy. In data-driven training, we do not know the equation and therefore can not calculate weights. Instead, we can estimate weights in a probabilistic way with the given dataset. Though, the estimated from the training data map \eqref{eq.em2} differs from the true map \eqref{eq.map_tm} calculated from the equations, it still represents the dynamics of the system and, more importantly, corresponds to some unknown ODE.

%--------------------------------------------------------
\begin{figure}
\vskip 0.2in
\begin{center}
\includegraphics[width=0.8\columnwidth]{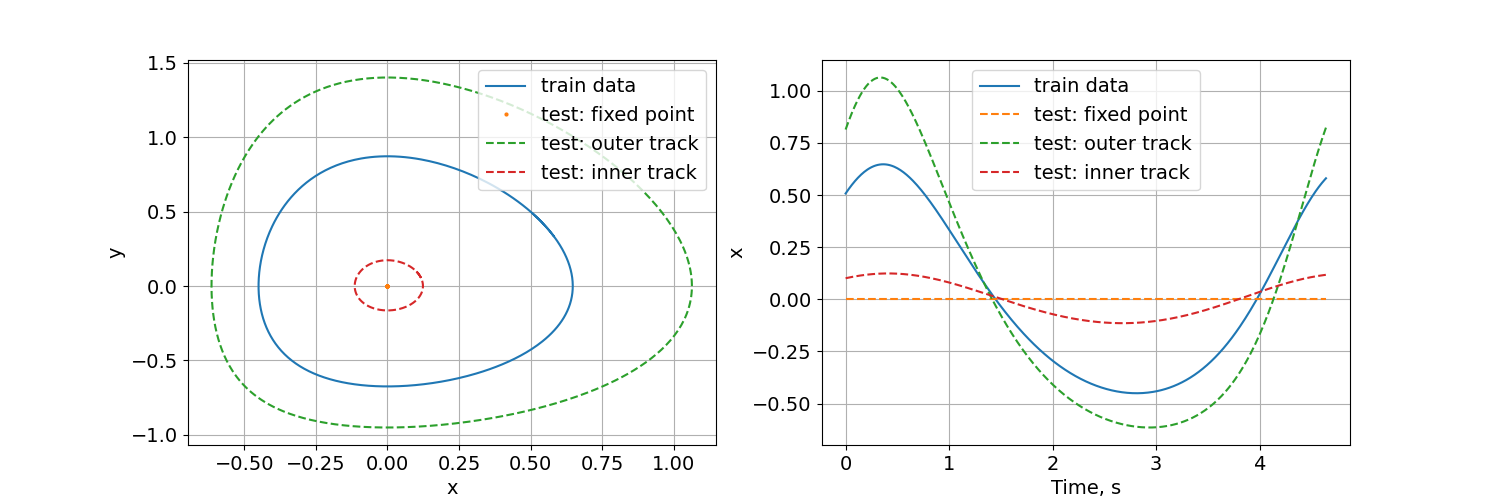}
\caption{Training (solid line) and testing unseen (dashed lines) data in phase space (left) and time space (right) for Lotka–Volterra system.}
\label{fig.07}
\end{center}
\vskip -0.2in
\end{figure}
%--------------------------------------------------------
\subsection{Identification of Lotka--Volterra system}
Let us now demonstrate that traditional neural networks are also difficult to apply for one-shot dynamics learning by an example of the Lotka--Volterra system
%--------------------------------------------------------
\begin{equation*}
\begin{aligned}
x' = y + xy,\;\;\;\;\;\;\;\;
y' = -2x - xy,
\end{aligned}
\end{equation*}
%--------------------------------------------------------
that describes the predator-prey population dynamics by nonlinear oscillations (see Fig.~\ref{fig.07}). Starting with an initial point $\XX_0 = (x_0, y_0)$, we calculate discrete states $\{\XX_i\}_{i=1,n}$ by numerical integrating of the differential equations with a constant time step. After data is generated, the system of ODEs is not used in further training.

We generate four different particular solutions that are presented as time series ${\XX(t_i)=(x_i, y_i)}$. As a training set, we use only one solution  starting at $(0.5, 0.5)$, while three other solutions with initial coordinates $(0.8, 0.8), (0.1, 0.1)$, and $(0.0, 0.0)$ are used for testing (see Fig.~\ref{fig.07}). For clarity, we generate solutions by integrating the equation from $t_0 = 0$ to $t = \unit[4.65]{s}$ with a constant time step $\Delta t = \unit[0.01]{s}$. The problem we address is following. Is it possible to recover dynamics of the whole system for unseen inputs knowing only a particular training solution. We consider three neural network architectures: proposed PNN with a third order map \eqref{tmap}, multilayer perceptron with sigmoid activation functions (MLP) and 5 hidden states, and long shot-term memory network (LSTM) with 5 inner cells.

After fitting NNs by the training particular solution, both MLP and LSTM networks tend to learn the given solution without any kind of generalization (see Fig.~\ref{fig.08}). They are not able to predict something that has not been presented in training data. At the same time, the proposed PNN predicts unknown dynamics in both nonlinear areas and near-linear oscillation around the stationary point. Moreover, it can even predict the stationary point without oscillation. Perhaps, it is possible to get the same level of generalization as for PNN by applying more intense training or slightly different settings of a state-of-the-art NN, but it is not clear how to achieve it.

For example, we focus  on LSTM architecture and try different parameters and training epochs. Namely, we vary the number of inner cells from 1 to 100 along with different regularization parameters and has not achieved the generalization of the dynamics from one sample. Fig.~\ref{fig.09} shows that LSTM either memorizes data or is under-fitted if the number of training epochs is too small or regularization terms are too large. The range of the regularization rate from 0.0 to 1.0 is scanned by the simple grid search procedure but it does not yield any improvement in generalization ability.

%--------------------------------------------------------
\begin{figure}[ht]
\vskip 0.2in
\begin{center}
\includegraphics[width=1.00\columnwidth]{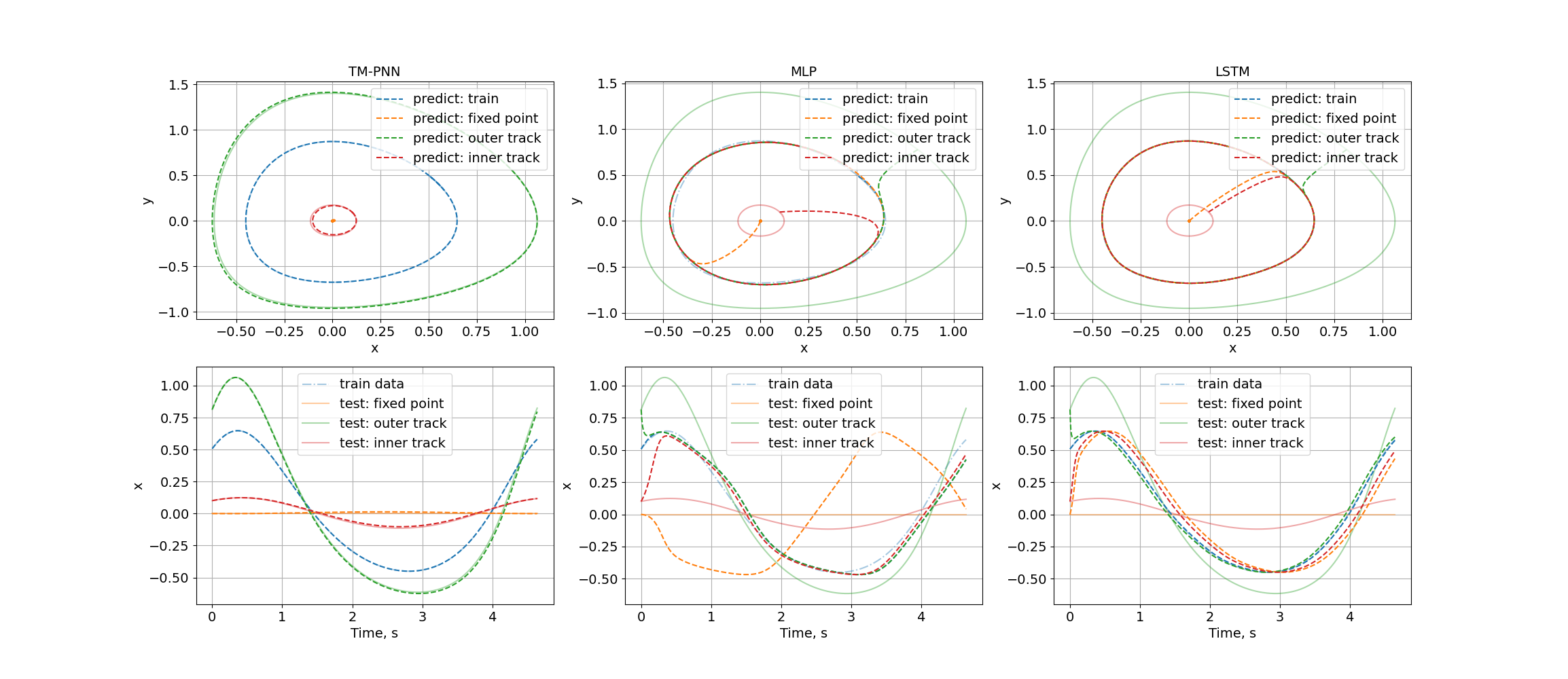}
\caption{Results of training and prediction for the new initial conditions by PNN (left), MLP (center), and LSTM (right). Dashed lines are predictions provided by the models. First row is a phase space, second row is a state space. PNN generalizes dynamics from one sample, while MLP and LSTM memorize the training sample.}
\label{fig.08}
\end{center}
%\vskip -0.2in
\end{figure}
%--------------------------------------------------------

%--------------------------------------------------------
\begin{table}
\caption{Training of LSTM with different hyper-parameters from one solution of Lotka–Volterra system}
\label{bb_train}
\vskip 0.15in
\begin{center}
\begin{small}
\begin{tabular}{|l|c|c|c|c|c|c|}
\toprule
& 1 inner cell & \multicolumn{3}{c}{5 inner cells}& 10 inner cells& 100 inner cells \\
Training&without& kernel& without& recurrent & without &without  \\
epochs&regularization& L1L2(0.1, 0.1)& regularization& L1L2(0.9, 0.9) & regularization &regularization  \\
\midrule
100 & underfitting& underfitting& underfitting& underfitting& underfitting & underfitting \\
1000 & underfitting& underfitting& memorization&underfitting& underfitting & underfitting \\
5000 & underfitting& underfitting& memorization& underfitting& memorization & underfitting \\
\bottomrule
\end{tabular}
\end{small}
\end{center}
\vskip -0.1in
\end{table}
%--------------------------------------------------------

%--------------------------------------------------------

%--------------------------------------------------------

Traditional neural networks have to be trained with lots of different solutions in order to perform well. It is still an open question if a state-of-the-art neural network can be trained  with only one solution of the dynamical system and achieve generalization for other ones. On the other hand, the Taylor map-based PNN is strongly associated with the theory of differential equations and is more suitable for dynamical systems learning. Fig.~\ref{fig.10} shows the mean squared error (MSE) between the true solutions and predictions provided by TM-PNN trained with one sample starting at initial conditions $(x_0=0.5, y_0=0.5)$. The increasing number of training epochs leads to decreasing MSE for unseen initial conditions.

This section demonstrates from both theoretical and practical points of view that the PNN is a more suitable architecture for dynamical systems learning rather than traditional ML models and neural networks. In this section, we consider virtual noise-free measurement just for demonstration of traditional ML and NN models limitations in the formulated problem of one-shot learning. The ODEs are used only for data generation, while weights of the PNN are estimated from the data without a priory knowledge of physical laws. On the other hand, if the system dynamics follows approximately a system of ODEs, the PNN can be initialized from these ODEs using the Taylor mapping approach (TM-PNN) and additionally fine-tuned with the data. The next sections consider this use case and correspond to real measurements with noisy and partial observations. We also do not provide a comparison with traditional ML and NN models due to their inapplicability for the problem of learning dynamical systems with one sample.

\begin{figure}[ht]
\vskip 0.2in
\begin{center}
\includegraphics[width=0.9\columnwidth]{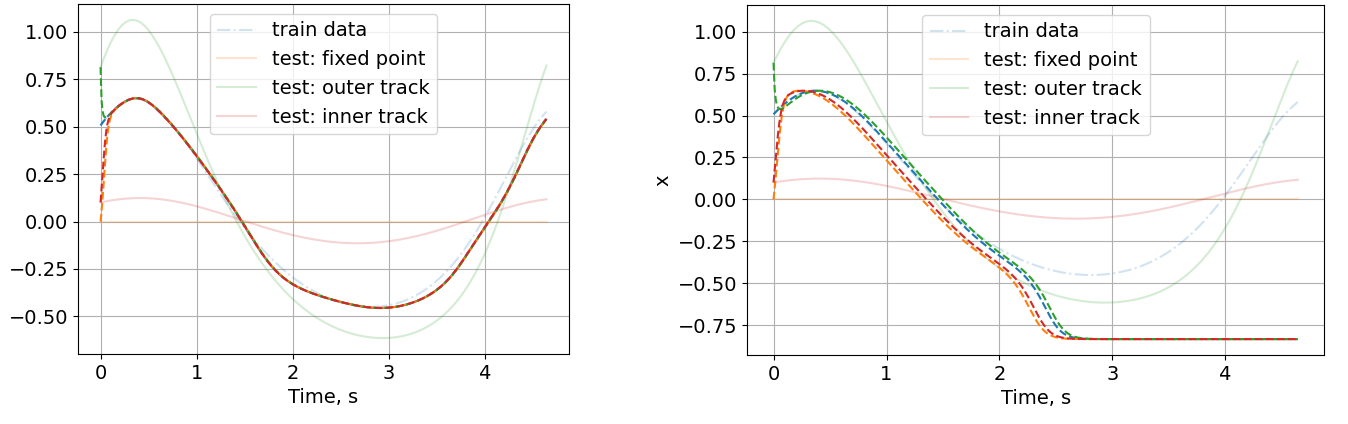}
\begin{minipage}[t]{0.49\textwidth}
\centerline{\small{a) memorization of the training solution}}
\end{minipage}
\begin{minipage}[t]{0.49\textwidth}
\centerline{\small{b) underfitting}}
\end{minipage}
\caption{Examples of memorization (a) and underfitting (b) of the LSTM with different hyper parameters.}
\label{fig.09}
\end{center}
\vskip -0.2in
\end{figure}

%--------------------------------------------------------
\begin{figure}[ht]
\vskip 0.2in
\begin{center}
\includegraphics[width=0.5\columnwidth]{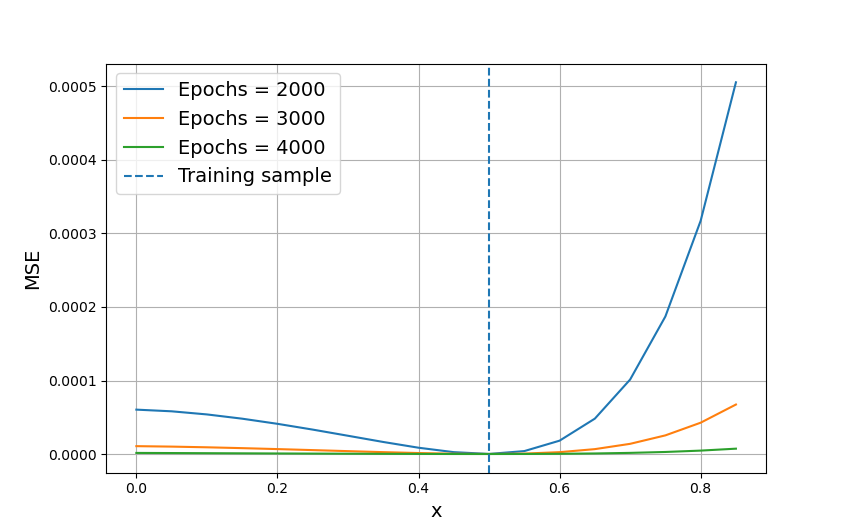}
\caption{Convergence of the TM-PNN during different number of training epochs for predictions with unseen inputs $x$ for one-shot learning of Lotka--Volterra system.}
\label{fig.10}
\end{center}
\vskip -0.2in
\end{figure}
%--------------------------------------------------------

\section{Fine-tuning of the real pendulum from idealized mathematical description}
\label{sec.4}

Let us explain the proposed approach with a simple example of a real pendulum. Having the measured oscillation of the pendulum for one initial angle, we would like to predict oscillations with new initial angles. In other words,
we aim at a generalized model of a real pendulum from only one observation. To measure data, we created a pendulum with targeted length $L=\unit[0.30]{m}$, but the true length became $L=\unit[0.28]{m}$ m because of some uncertainties. Instead of operating with the exact length, we consider the initial assumption $L=\unit[0.30]{m}$ as only available a priori knowledge about the system.

To measure the pendulum oscillations, we recorded video streams of $\unit[5]{s}$ with $\unit[10]{fps}$. To introduce additional noise, the camera is not calibrated, and the measurements are not filtered. The angle $\phi_i$ of the pendulum is estimated in each $i$th video frame, while the angular velocity $\phi'_i$ remains non-observable. In this way, each sample of the pendulum oscillation during $\unit[5]{s}$ is represented by time series $\{\phi_i\}$ for $i=0..49$ with time step $\Delta t=\unit[0.1]{s}$. The oscillations are damped, which leads to oscillation amplitude decay. 

\subsection{Translating the ODE of ideal pendulum into a Taylor map}
The simple physics-based model of the pendulum can be described by the differential equation $\phi'' = -g\cdot \sin(\phi)/L$, where $'$ is derivative on time, $\phi$ is the angle of the pendulum, and $g,\,L$ are parameters. This equation can be written in a matrix form up to the third order nonlinearities:
%--------------------------------------------------------
\begin{equation}
    \label{eq:pend}
    \frac{d}{dt}
    \begin{pmatrix}
        \phi\\
        \phi'\\
    \end{pmatrix}=
        \begin{pmatrix}
        0&1\\
        -g/L&0
        \end{pmatrix}
        \begin{pmatrix}
        \phi\\
        \phi'
        \end{pmatrix}+
        \begin{pmatrix}
        0&0&0&0\\
        g/(6L)&0&0&0
        \end{pmatrix}
        \begin{pmatrix}
        \phi^3\\
        \phi^2\phi'\\
        \phi\phi^{'2}\\
        \phi^{'3}
    \end{pmatrix}=
        P_1
        \begin{pmatrix}
        \phi\\
        \phi'
        \end{pmatrix}+
        P_3
        \begin{pmatrix}
        \phi^3\\
        \phi^2\phi'\\
        \phi\phi^{'2}\\
        \phi^{'3}
        \end{pmatrix}.
\end{equation}
%--------------------------------------------------------

The mathematical pendulum \eqref{eq:pend} theoretically continues the oscillation with the same amplitude and frequency indefinitely. Though the behaviour differs from the real damped oscillation, this simplified physics-based model can still initialize the PNN with some level of accuracy. Let us calculate a Taylor map for the time step $\Delta t = \unit[0.1]{s}$ following the algorithm presented in \cite{TMPNN}:
%--------------------------------------------------------
\begin{equation}
	\label{tmap_cd}
	\MM(t): 
	\begin{pmatrix}
	\phi\\
	\phi'
	\end{pmatrix}
	= W_1\,
	\begin{pmatrix}
	\phi_0\\
	\phi'_0
	\end{pmatrix}
	+W_2\,
	\begin{pmatrix}
	\phi_0^2\\
	\phi_0\phi'_0\\
	\phi_0^{'2}
	\end{pmatrix}+
	W_3
	\begin{pmatrix}
    \phi_0^3\\
    \phi_0^2\phi'_0\\
    \phi_0\phi_0^{'2}\\
    \phi_0^{'3}
    \end{pmatrix}.
\end{equation}
%--------------------------------------------------------
By denoting $\XX = (\phi,\,\phi'),\;\XX_0=(\phi_0,\,\phi'_0)$ and substituting \eqref{tmap_cd} to \eqref{eq:pend}, one can write
%--------------------------------------------------------
\begin{equation}
\begin{split}
\XX'
=P_1W_1\XX_0+P_1W_2\XX_0^{[2]}+\left(P_1W_3 + P_3W_1^{[3]}\right)\XX_0^{[3]}
+
\mathcal{O}(\XX_0^{[3]}),
\end{split}
\label{tmap_cd2}
\end{equation}
where $W_1^{[3]}$ can be calculated from the relation $(W\XX)^{[3]} = W^{[3]}\XX^{[3]}$. For instance, for $W = \{w_{ij}\}$,
\begin{equation*}
\begin{split}
&(W\XX)^{[3]}=
\begin{pmatrix}
w_{11}\phi+w_{12}\phi'\\
w_{21}\phi+w_{22}\phi'\\
\end{pmatrix}^{[3]}=\\
&\begin{pmatrix}
w_{11}^3&3w_{11}^2w_{12}&3w_{11}w_{12}^2&w_{11}^3\\
w_{11}^2w_{21}& w_{11}^2w_{22} + 2w_{11}w_{12}w_{21}&w_{12}^2w_{21} + 2w_{11}w_{12}w_{22}&w_{12}^2w_{22}\\
w_{21}^2w_{11}& w_{21}^2w_{12} + 2w_{21}w_{22}w_{11}&w_{22}^2w_{11} + 2w_{21}w_{22}w_{12}&w_{22}^2w_{12}\\
w_{21}^3&3w_{21}^2w_{22}&3w_{21}w_{22}^2&w_{22}^3\\
\end{pmatrix}
\begin{pmatrix}
\phi^3\\
\phi^2\phi'\\
\phi\phi^{'2}\\
\phi^{'3}
\end{pmatrix}
=W^{[3]}\XX^{[3]}.
\end{split}
\end{equation*}
%--------------------------------------------------------
Taking the derivative of \eqref{tmap_cd} and comparing it with \eqref{tmap_cd2}, one can obtain a system of ODEs that do not depend on $\XX_0$  and represent the dynamics of matrices $W_1 = W_1(t), W_2 = W_2(t)$, and $W_3 = W_3(t)$:
%--------------------------------------------------------
\begin{equation}
\begin{split}
W'_1 =P_1W_1,\;\;\;\;
W'_2 =P_1W_2,\;\;\;\;
W'_3 =P_1W_3 +P_3W_1^{[3]}.
\end{split}
\label{tmap_cd3}
\end{equation}
%--------------------------------------------------------
Solving \eqref{tmap_cd3} for the time interval $[0; 0.1]$ with the initial conditions $W_1(0) = I, W_2(0) = 0, W_3(0) = 0$ and $I$ as the identity matrix results in a Taylor map that describes the dynamics of the ideal pendulum during $\Delta t = 0.1$. For instance, for $g=9.8$ and $L=0.3$, the solution of \eqref{tmap_cd3} up to the two digits is
%--------------------------------------------------------
\begin{equation}
        W_1=
    \begin{pmatrix}
        0.84&0.09\\
        -3.1&0.84\\
    \end{pmatrix},\;\;\;\;W_2 = 0,\;\;\;\;
    W_3=
    \begin{pmatrix}
        0.02&0.0023&0.00012&2.3\cdot10^{-6}\\
        0.43&0.064&0.0044&0.00012\\
    \end{pmatrix}.
    \label{eq:initmap}
\end{equation}
%--------------------------------------------------------

%--------------------------------------------------------
\begin{figure}
\vskip 0.2in
\begin{center}
\centerline{\includegraphics[width=0.6\columnwidth]{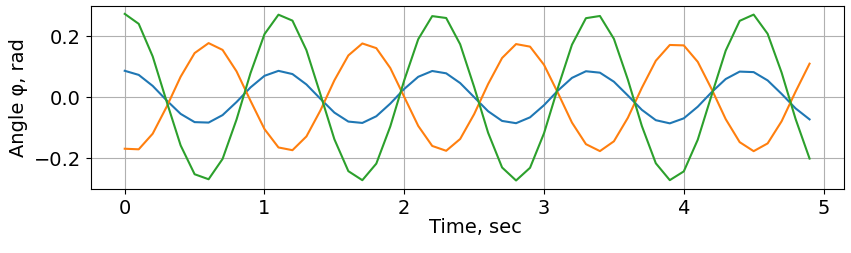}}
\caption{Predictions of initialized from ODE \eqref{eq:pend} TM-PNN for different initial angles. The TM-PNN initially represents a rough assumtion about the pendulum dynamics.}
\label{fig.11}
\end{center}
\vskip -0.2in
\end{figure}
%--------------------------------------------------------

Fig.~\ref{fig.11} shows that map \eqref{tmap_cd} with weights \eqref{eq:initmap} represents a theoretical oscillation of the mathematical pendulum with $L=\unit[0.30]{m}$. Though the initialized with \eqref{eq:initmap} PNN only roughly approximates the real pendulum, it can be used for the physical prediction of the dynamics starting with arbitrary angles.

%--------------------------------------------------------
\begin{figure}[ht]
\vskip 0.2in
\begin{center}
\centerline{\includegraphics[width=0.5\columnwidth]{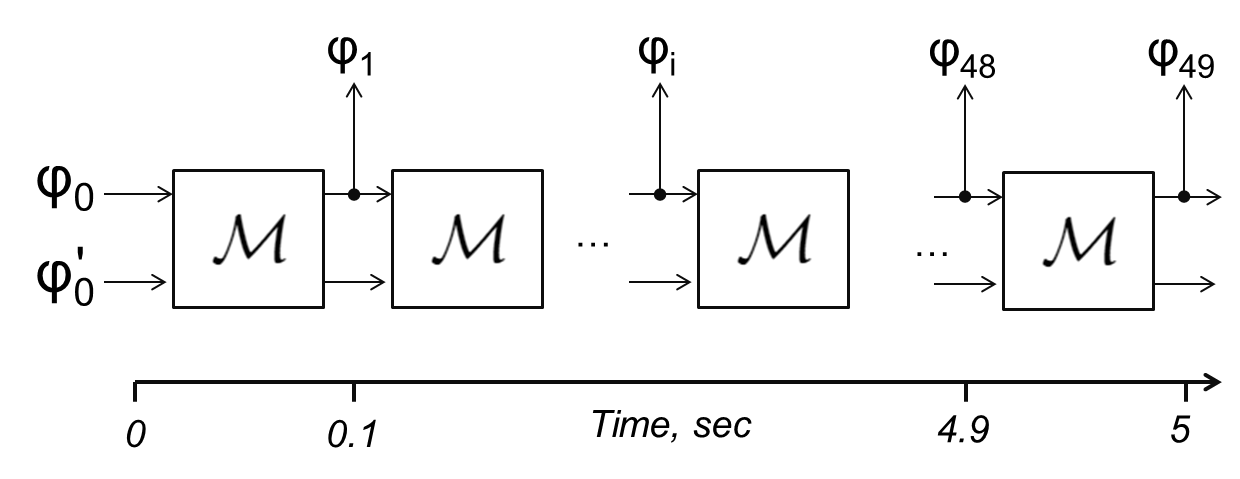}}
\caption{Multi-output TM-PNN architecture with shared weights and partial observations.}
\label{fig.12}
\end{center}
\vskip -0.2in
\end{figure}
%--------------------------------------------------------
\subsection{One-shot learning of the real pendulum from one observation}

%The equation \eqref{eq:pend} and corresponded Taylor map \eqref{eq:initmap} represent a theoretical oscillation of the mathematical pendulum. The pendulum will theoretically continue oscillation with the same amplitude and frequency indefinitely. For a real pendulum, the length $L$ may differ from the assumption of $L=0.3$ and there should be a damping effect that leads to the oscillation amplitude decay. Also, let us assume that one can measure only angles $\phi$ while angular velocities $\phi'$ are not observable.

Instead of fitting the ODE for the real pendulum, we fine-tune the Taylor map--based PNN (TM-PNN).
Since angles are measured every $\unit[0.1]{s}$ for $\unit[5]{s}$ in total, we constructed a Taylor map--based PNN (TM-PNN) with 49 layers with shared weights $\mathcal M$ initialized with \eqref{eq:initmap}. Since the angular velocities $\phi'_i$ are not observable, the loss function is MSE only for angles $\phi_i$. The TM-PNN propagates the initial angle $\phi_0$ along the layers and recovers the angular velocities as latent variables.

\begin{figure}[ht]
  \centering
  \includegraphics[width=0.9\columnwidth]{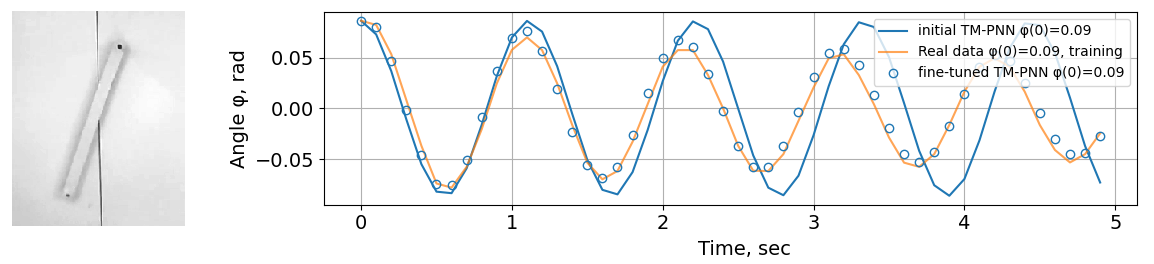}
  \caption{One-shot tuning of the TM-PNN for real pendulum with initial weights obtained from the theoretical ideal ODEs.}
  \label{fig.13}
\end{figure}

The TM-PNN presented in Fig.~\ref{fig.12} is implemented as a multi-output model in Keras with a TensorFlow backend. The Adam optimizer with a controlled gradient clipping during 1000 epochs is used for training. The TM-PNN with initial weight \eqref{eq:initmap} is fine-tuned with one oscillation of the real pendulum with initial angle $\phi_0=0.09, \phi'_0=0$. Fig.~\ref{fig.13} shows the oscillation provided by the initial TM-PNN \eqref{eq:initmap} with a blue solid curve. This infinite oscillation represents the theoretical assumption on the dynamics of the pendulum with length $L=0.30$. The orange solid line represents a real pendulum oscillation with true length $L=0.28$ and the damping of the amplitude that is used for training. The prediction of the fine-tuned TM-PNN is presented by the blue circles. 

%--------------------------------------------------------
\begin{figure}
  \centering
  \includegraphics[width=0.9\columnwidth]{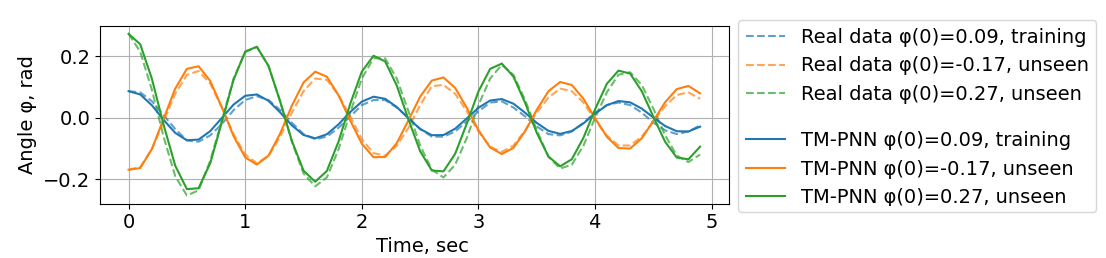}
  \caption{Predictions of the fine-tuned TM-PNN with unseen initial angles.}
  \label{fig.14}
\end{figure}

%--------------------------------------------------------
The fine-tuning of the TM-PNN with one oscillation not only increases the accuracy of the prediction for the given training oscillation but also, more importantly, preserves physical consistency for the predictions starting with unseen angles. 
Fig.~\ref{fig.14} compares the predictions provided by the fine-tuned TM-PNN for unseen angles with measurements of real pendulum oscillation. As a result, we have a TM-PNN model that has been trained only with one initial angle and predicts the dynamics for other angles.

%--------------------------------------------------------
%--------------------------------------------------------

\section{One-shot recovering of complex physics in charged particle accelerators}
\label{sec.5}
Since Taylor mapping is commonly used in accelerator physics \cite{ref13,SLAC-PUB-9574}, demonstrating  the advantages of the proposed TM-PNN is beneficial in this field. In this section, the deep TM-PNN is constructed to recover the complex physics of one of the largest worldwide X-ray sources, PETRAIII \cite{article}. We initialize a deep TM-PNN with the theoretical ODEs that describe the dynamics of particles and then fine-tune the TM-PNN with noisy, limited, and partial observation from the real machine.

\subsection{Problem formulation}
The PETRAIII storage ring consists of 1519 magnets and provides the transportation of the electron beam along the 2.3 km of ring length. For simplicity, we consider the particle motion only in horizontal $(x,x')$ and vertical $(y,y')$ planes without considering energy deviation. Thus, the state vector $\XX = (x,x',y,y')$ represents the location and velocities of a particle beam, and is propagated consequentially through all the magnets.

The circular particle accelerator $\mathcal A$ transfers the particle beam with initial coordinates $\XX(0)=\XX_0$ at the beginning of the ring to the state $\XX(1)$ at the end of the ring during the first turn. The multi-turn dynamics is represented by consequentially transferring the beam with the coordinates received at the previous turn. For example, for $n$ turns, one can write
\begin{equation}
    \XX(n) = \underbrace{\mathcal A \circ \mathcal A \circ \ldots \circ \mathcal A}_{n}  \circ \XX_0.
    \label{eq:A}
\end{equation}
One of the most important characteristics of the charged particles motion is the multi-turn frequencies of the oscillation. Having the beam coordinates at each turn $\XX(i)$, the main frequencies of the multi-turn oscillation in the horizontal and vertical planes can be calculated. Since these main frequencies have an important role in accelerator design and can be considered as an operational regime, it is important to know the true frequencies in the real accelerator.

The dynamics of the beam in the real accelerator differs from that in the theoretical design given lots of imperfections in the construction and operation conditions. For example, Fig.~\ref{fig.07} represents the theoretical and real beam tracks during one turn of the ring in the experiment in PETRAIII. In the example, we demonstrate how the proposed approach can be used to recover the multi-turn dynamics of the real accelerator with the help of one measured beam track from only the first turn.

The measurements of the beam during the first turn define one training sample. A total of 246 beam position monitors (BPMs) measure the $(x,y)$ locations of the beam along the ring. So the training sample is represented by a time series of two variables $\{x_i, y_i\}$ with  246 stamps that define the first turn of the beam around the ring (Fig.~\ref{fig.07}).

The main idea of the approach is to train a TM-PNN $: \XX(0)\rightarrow \XX(1)$ with one training sample and estimate multi-turn frequencies by replacing the measurements from the real accelerator $\mathcal A$ in \eqref{eq:A} with the predictive model. This means that the TM-PNN has to provide accurate predictions not only for a single initial condition $\XX(0)=\XX_0$ but also for $n$ new coordinates of the beam at each turn.

\begin{figure}
  \centering
  \includegraphics[width=5.5in]{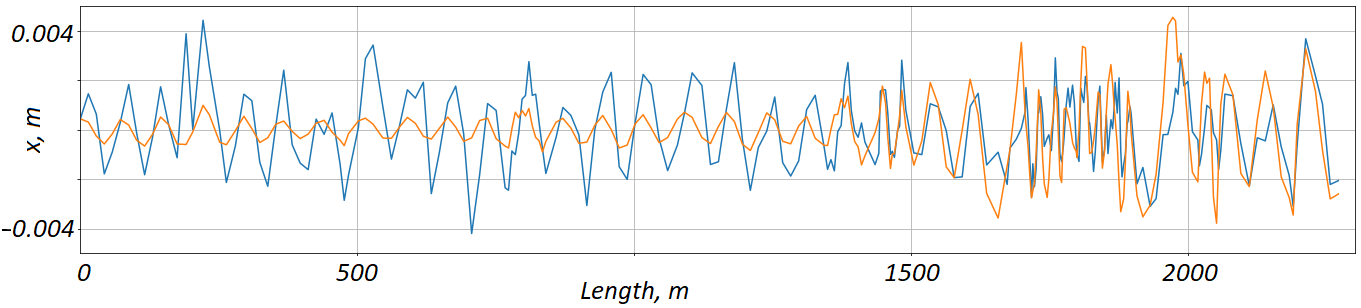}
  \caption{Theoretical one-turn beam trajectory (orange line) and measured trajectory (blue line). Each trajectory is represented by 246 stamps where BPMs are located.}
  \label{fig.07}
\end{figure}

\subsection{TM-PNN architecture of the particle accelerator}
Each of the 1519 magnets is described by a system of ODEs. For instance, for motion in the horizontal plane, one can write an equation in a general form \cite{ref13}
\begin{equation}
x'' = \frac{qH}{m_0\gamma v}\left(-(1 + x^{'2})B_y + y'(x'B_x + B_s)\right),
\label{eq.phb}
\end{equation}
where $x$ means the particle location, $x'$ is derivative with respect to the length along the lattice, $(B_x, B_y, B_s)$ represents the magnetic field, $q, m_0$ are parameters, and $H, \gamma, v$ are functions of $(x, x')$. 

To represent these ODEs as Taylor maps, we limited  ourselves to the second order of nonlinearities and built 1519 Taylor maps for each magnet with the help of the OCELOT framework \cite{Agapov:2014yku}. The architecture of the TM-PNN is presented in Fig.~\ref{fig.08}. There are 1519 layers with unique weights that are not shared. Since there are 246 BPMs located along the ring, the TM-PNN has 246 outputs. Each output represents beam location ($x,y$) in the horizontal and vertical planes; velocities $(x', y')$ are not observable and considered as latent variables. The initialized from ODEs TM-PNN accurately represents the theoretical assumption of the beam dynamics.

\subsection{One-shot tuning of the TM-PNN}
To represent system uncertainty in the experiments, we decreased the strength of only one of the 1519 magnets by 20\% and measured one beam trajectory for the first turn (see Fig.~\ref{fig.07}). This  trajectory represents a time series $\{x_i, y_i\}$ with 246 stamps for BPM measurements of the first turn around the ring. For fine-tuning the TM-PNN, we use the following loss function:
\begin{equation}
\label{eq:loss}
    Loss =\sum_{i=0}^{246}||\XX(0)_i^\text{TM-PNN} - \XX(0)_i^\text{BPM}||+ \lambda \sum_{j=0}^{1519}S(W_1^j, W_2^j),
\end{equation}
where training data $\XX(0)_i^{\text{BPM}}$ is the measurement of the $i$th BPM in the first turn, $\XX(0)_i^\text{TM-PNN}$ is the $i$th output of the TM-PNN  with input $\XX(0)=\XX_0$ from the training data, $\lambda = 1e-10$ is the rate, and $S$ is the symplectic penalty for each layer that is defined by the symplectic property.

The symplectic property \cite{arnold1989mathematical} is an essential invariant that has to be preserved for the physical consistency of the Hamiltonian system. Since the particle motion can be represented by Hamiltonian dynamics, the symplecticity of each hidden layer $\mathcal M_i: \XX_{i-1} \rightarrow \XX_i$ has to be preserved during training:
\begin{equation}
\label{eq:sympl}
    \left(\frac{\partial \XX_i}{\partial \XX_{i-1}}\right)^T\,J\;\frac{\partial \XX_i}{\partial \XX_{i-1}} - J = 0,\;\forall \XX_{i-1},\;J = \begin{pmatrix}0&I\\-I&0\end{pmatrix},
\end{equation}
where $I$ is an identity matrix and $T$ means the transpose. The symplectic property \eqref{eq:sympl} for the TM-PNN leads to algebraic constraints on weights $W_i = \{w_i^{jk}\}$ that do not depend on $\XX_{i-1}$. It guarantees the physical property of the trained model whatever inputs $\XX_0$ are used. For example, for the second-order Taylor map with $W_1=\{w_1^{ij}\},\;W_2=\{w_2^{mn}\}$ where $i,m$ represents the indices of the rows and $j,n$ those of the columns, the condition \eqref{eq:sympl} yields constraints
\begin{equation}
\label{eq:algsympl}
    \begin{split}
    &w_1^{11}w_1^{22} - w_1^{12}w_1^{21} - 1 = 0,\;\;\;w_2^{11}w_2^{23}-w_2^{13}w_2^{21} = 0,\;\;\;w_2^{12}w_2^{23} - w_2^{13}w_2^{22} = 0,\\
    &w_1^{11}w_2^{22}-w_1^{21}w_2^{12} + 2w_1^{22}w_2^{11} - 2w_1^{12}w_2^{21} = 0,\\
    &w_1^{22}w_2^{12}-w_1^{12}w_2^{22} + 2w_1^{11}w_2^{23} - 2w_1^{21}w_2^{13} = 0,\\
    \end{split}
\end{equation}
with the penalty $S$ as the sum of squares of all left-hand terms in \eqref{eq:algsympl}. Since this penalty does not depend on the inputs, the physical structure of the layers is preserved for all new inputs, which has a large impact on generalization. If the symplectic regularization is not considered during the training or traditional nonphysical  L1L2 regularization is used, the tuning of the maps leads to the overfitting of the model, which causes nonphysical  predictions.

\begin{figure}
  \centering
\begin{minipage}[t]{0.61\textwidth}
  \includegraphics[width=2.5in]{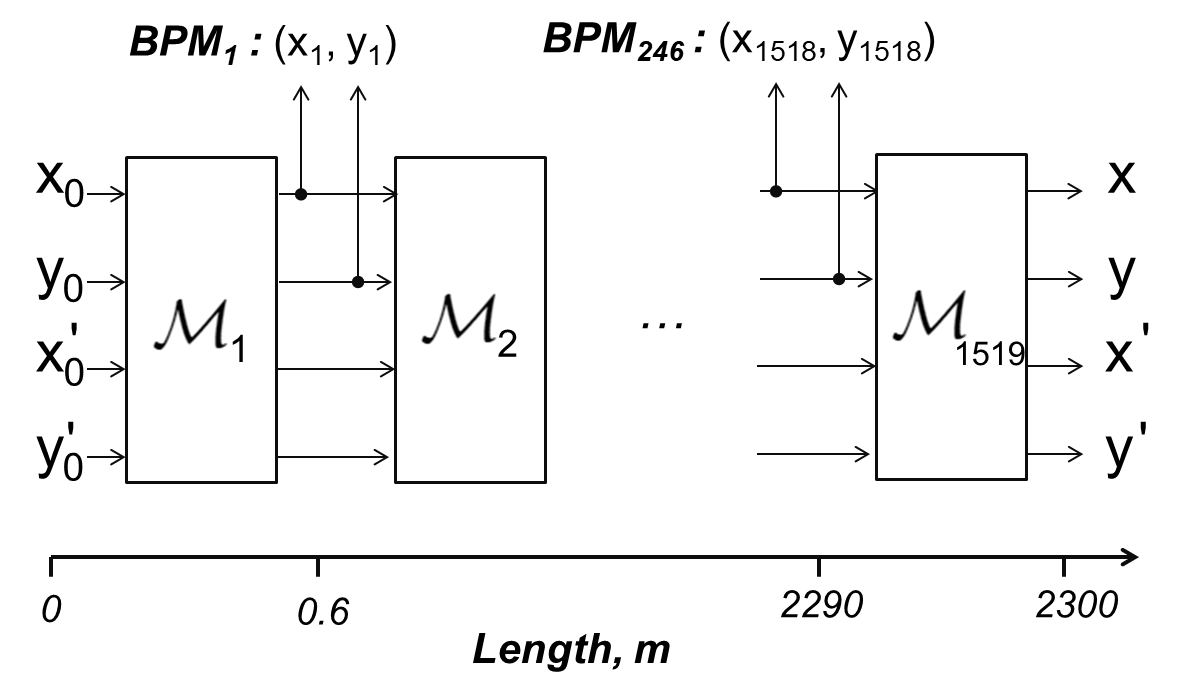}
  %\caption{Theoretical beam trajectory (orange line)\hspace{1.5cm}}
  %and measured trajectory (blue line).
  \caption{Multi-output deep TM-PNN for the \hspace{2.5cm}}
   PETRAIII storage ring.
  \label{fig.08}
    \end{minipage}
    %\hspace*{0.5cm}
    \begin{minipage}[t]{0.38\textwidth}
      \includegraphics[width=2.0in]{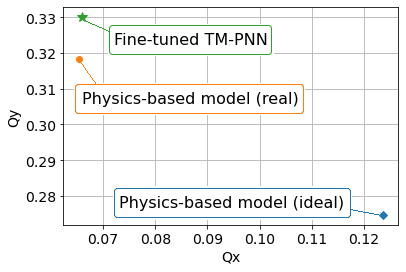}
  \caption{Frequencies in the horizontal (Qx) and vertical (Qy) planes.}
  \label{fig:07}
    \end{minipage}
\end{figure}

\subsection{Physics recovering with the fine-tuned TM-PNN}
The fine-tuned TM-PNN accurately represents the real beam track during the first turn with the one initial beam coordinates $\XX_0$ and preserves the physical consistency of the predictions for arbitrary inputs via symplectic regularization. So, the TM-PNN can simulate multi-turn dynamics in the accelerator by replacing $\mathcal A$ in \eqref{eq:A} with the TM-PNN model and predict the main oscillation frequencies.

Since we know exactly which magnet is affected, we can estimate the main frequencies with the physics-based model based on Equation \eqref{eq.phb}. Fig.~\ref{fig:07} shows that the true frequencies calculated in the OCELOT  coincide with the TM-PNN prediction with fair accuracy. The main horizontal frequency predicted by the TM-PNN in 500 virtual turns has a relative error less than 1\%, and the vertical one has an error less than 5\%. Note that to calculate true frequencies, one has to know exactly which magnet was affected. In real operation conditions, this information is not available, but the fine-tuned TM-PNN recovers physical properties from partial, noisy, and limited observations from the accelerator.

\section{The representation capacity of the TM-PNN}
\label{sec.7}
Though the presented approach has some limitations and relies mostly on the specific task of one-shot learning of complex dynamical systems, it can be widely used for dynamical system identification. In science and engineering, there are tasks when it is impossible, for some reason, to collect or simulate enough data to train black-box models. Also, the presented approach targets situations when the physics-based and gray-box models are ineffective in the computational sense given the complexity of the considered systems. Otherwise, parametrizing the system of ODEs and estimating the parameters with statistical methods or gray-box modeling would be easier.

Since the Taylor maps \eqref{tmap} entail the calculation of Kronecker products, limitations are observed  in the scalability of the direct application of the technique with extremely high orders. Further research on this topic should be done but hopefully can be based on the existing works. For example, in \cite{Yu}, a tensor-train decomposition is adopted to learn low-dimensional representations of the higher-order weight tensors obtained from Kronecker products, while the authors in \cite{ref201} suggest a stochastic Riemannian optimization procedure to train models based on Kronecker products.

On the other hand, the presented technique can be directly adopted for physical systems when high orders have not arisen or have neglected influence. For example, third-order nonlinearities along with a deep architecture of one thousand layers are often enough in charged particle accelerators. Moreover, even the complex behavior of the dynamical chaos can be described by the second-order polynomial map \cite{Henon}.

\subsection{Dynamical systems with polynomial ODEs of low orderes}

The systems of ODEs with polynomial time-independent right-hand sides are widely used in science and engineering. For example, in \cite{6MP}, the authors consider modelling the cell metabolism of 6-mercaptopurine, one of the most important chemotherapy drugs. The paper introduces the system of ODEs with ten equations with polynomial nonlinearities up to the third order. The authors indicate that the physics-based model is overcomplicated and requires a knowledge of multiple kinetic parameters. They suggest considering a Boolean network instead but point to its over-simplicity at the same time. Since the TM-PNN has the same representation capacity as ODEs and is represented simply by weights, it can potentially solve this complexity-accuracy trade-off.

The paper \cite{AC} presents the system of ODEs for stability analysis and the control of the nonlinear dynamics of an articulated car-trailer system. The authors point out possible instabilities in motion and control. The TM-PNN can be used for predictive control in real operation conditions. After initializing from the equations of vehicle motion that are known by design, the TM-PNN can be fine-tuned with sensor-based observations continuously in time and guarantee the physical consistency of the model. The symplectic regularization is also applicable to this example of a Hamiltonian system.

\subsection{Cavitation as an example of non-polynomial and time-dependent ODEs}

This example briefly demonstrates the application of the TM-PNN for dynamical systems that are described by the ODEs with  non-polynomial and time-dependent nonlinear right-hand sides. Cavitation is the formation of gas or vapor bubbles in a liquid \cite{cavitation1989}. The growth, collapse, and rebound of a cavitation bubble traveling along the flow is governed by the Rayleigh–Plesset equation:
%--------------------------------------------------------
\begin{equation}
\label{eq.RP_full}
\rho\left(R\ddot R+ \frac{3}{2}\dot R^2\right)=p_B-p_0+p_a\sin(\omega t)-\frac{2\sigma}{R}-\frac{4\mu}{R}\dot{R},
\end{equation}
%-------------------------------------------------------
where $R$ is the bubble radius, $\dot R$ is the derivative on time, $\sigma$ is surface tension, $\mu$ is viscosity, $\omega$ is the driving frequency, and $\rho$ is the density of the liquid. For simplicity, we consider $p_B, p_0$, and $p_a$ as constant parameters. Though Equation \eqref{eq.RP_full} contains non-polynomial nonlinear functions and even depends on time directly, it can be represented in the form that allows one to apply the Taylor mapping technique. Indeed, after the introduction of new variables $x = R,\; y = \dot R,\;z=1/R,\;s =\sin(wt),\;c =\cos(wt)$, Equation \eqref{eq.RP_full} can be presented in the polynomial form with a five-dimensional state vector:
%--------------------------------------------------------
\begin{equation}
\begin{split}
\dot x &= y,\;\;\;\;\;\;\dot z = -yz^2,\;\;\;\;\;\;\dot s = \omega c,\;\;\;\;\;\;\dot c = -\omega c,\\
\dot y &= -1.5y^2z + z(p_B-p_0)\rho+(zsp_a-2\sigma z^2-4\mu yz^2)/\rho.\\   
\end{split}
\label{eq:RPP}
\end{equation}
%--------------------------------------------------------
The system \eqref{eq:RPP} is equivalent to \eqref{eq.RP_full} and allows the translation to TM-PNN directly. New variables play a role of new features that should be additionally constructed and incorporated into the TM-PNN. This means that given an oscillation from experimental analysis, the TM-PNN can be used for identifying a true model that represents the real bubble oscillation for new unseen conditions.

%\section{Code}

\section{Results and Further Work}

Since Taylor maps can be used for solving systems of ODEs with the necessary level of accuracy, the TM-PNN model is suitable for solving inverse problems of system identification from the measured data. We firstly demonstrate that TM-PNN can successfully recover the general solution of the ODE from a particular solution with examples of free fall and  Lotka--Volterra oscillation. For the problem of a body free fall, we demonstrate that surrogates models require lots of data to represent the dynamics of the system. The existing and well-developed NN models are also difficult to apply to recover physics from one solution of the dynamical system. In both cases, traditional and state-of-the-art ML and NN models provide unsatisfactory performance for unseen input beyond the training data.  For the considered examples, traditional ML and NN are difficult to apply while the proposed TM-PNN can be trained from scratch with limited data.

On the other hand, if the dynamics of a system follows approximately a given differential equation, the Taylor mapping technique can be used to initialize the weights of the TM-PNN. This allows fine-tuning of the model from one training sample of a real system dynamics. We demonstrate in practice for the real pendulum and X-ray source that the proposed technique allows recovering the physical properties of the systems from noisy and partial observations. In these examples, we use prior but inaccurate physical knowledge about the system to initialize TM-PNN. This initial approach of weights is further fine-tuned from one measurement to represent the real system dynamics.

The symplectic regularization is suggested for Hamiltonian systems. The symplectic property is utilized to preserve the physical properties of the TM-PNN during training. The considered regularization  penalties significantly differ from the traditional L1 or L2 norm that are used in the ML field. Further comparison and investigation on this topic are required for the learning of dynamical systems.

One of the directions in the development of the proposed model is its utilization for the automatic derivation of the physics-based model. The key point is that the TM-PNN tuned with data corresponds to some unknown system of ODEs. If it is possible to translate TM-PNN to ODEs, this can result in a new physics-based model. Using the described in \eqref{tmap}-\eqref{eq:dW} algorithm, one can potentially solve a boundary problem and identify the system of ODEs that should be approximately equivalent to Taylor maps extracted from the trained TM-PNN. The translation of a system of ODEs into the TM-PNN implies truncation of a map \eqref{tmap} and requires solving a new differential equation \eqref{eq:dW} for the weight. In contrast to this, the inverse problem will probably consist of an integral equation along with the truncation of the right-hand side of the ODEs system \eqref{odesystem}.

\section{Conclusion}

The paper proposes an approach to incorporate physics-based models into the TM-PNN architecture. This allows the preservation of a priori, physical knowledge in the NN and fine-tuning it with one sample. The physics-based structure of the TM-PNN also provides the possibility of easily introducing a physical constraint for the model, which was demonstrated with the example of symplectic regularization.

If nothing is known about the dynamical system in terms of even approximate ODEs, the option of collecting data and training a state-of-the-art black-box model from large data sets is available. The paper does not compare the proposed approach with such methods and NN architectures because of differences in the problem formulations. Also, while training a state-of-the-art NN model with only one time series of 246 stamps of particle accelerator dynamics and achieving physically accurate predictions for long-term dynamics for 500 new unseen time series are possible in theory, such tasks were not solved before in the community and require additional efforts in a separate study.

In contrast to this, the huge amount of physics-based models in the form of ODEs have been developed over the years and describe processes in mechanics, robotics, thermodynamics, fluid mechanics, and other fields. The paper discusses a clear approach for transferring these physics-based models to the NN and avoiding time-consuming numerical solvers and big data sets for training. To process noisy data, the multi-step architecture of the TM-PNN is used. This means that the input of the TM-PNN is only an initial state vector, while dynamics in time is predicted based on previous predictions. %Also, physics-based regularization helps avoid overfitting when the TM-PNN is trained with measured data. 

Based on the theoretical equivalence of the ODE systems and the TM-PNN, the paper demonstrates in practice that fine-tuning the TM-PNN from one sample works not only for a simple pendulum but also for a complex particle accelerator with noisy, limited, and partial observations. The example on cavitation demonstrates that the TM-PNN architecture can be widely applied for physical systems with non-polynomial and time-dependent ODEs in areas other than accelerator physics.

Since this is the first time the connection between ODEs and the TM-PNN is presented in terms of the one-shot learning of dynamical systems, further research on the estimation of accuracy, performance, and limitations of the proposed method should be conducted. Also, the symplectic regularization for the Hamiltonian systems should be investigated in more detail and compared with other regularization penalties as this may be helpful for solving and speeding up real physics problems. How accurate the initial assumption of the system of ODEs and the complex TM-PNN architecture is in terms of nonlinear orders or the number of hidden layers required for a given physical problem can also be openly questioned.

\bibliographystyle{unsrt}  
\bibliography{paper}

\end{document}